# Simple method to eliminate blur based on Lane and Bates algorithm

S. Aogaki, I. Moritani, T. Sugai, F. Takeutchi, and F.M. Toyama, Kyoto Sangyo University, Kyoto-603, Japan

*Abstract*-- A simple search method for finding a blur convolved in a given image is presented. The method can be easily extended to a large blur. The method has been experimentally tested with a model blurred image.

*Index Terms*—Deconvolution, Image restoration,

## I. INTRODUCTION

The blind deconvolution introduced by Lane and Bates (LB) is an innovative method in image restoration [1]. The method enables to uniquely remove blurs convolved in a given image. However, due to its extremely advanced analytical character, the computational complexity of the image processing is large. Therefore, it is meaningful to devise a technical tool to reduce the computational complexity.

The LB's method is based on the zero-sheets of the $z$-transform of a given image. In our preceding paper [2] we presented conditional expressions (CEs) that make it possible to automatically find zero-values for blurs convolved in the image. The CEs are very useful in detecting blurs almost automatically. However, when the size of blurs is large, it is not easy to find an appropriate expression for CEs, and also the computational load becomes immediately very large.

In this paper we present a search algorithm for finding blurs, which is a new mathematical tool for the LB's method but different from the CEs. Although the amount of computation becomes larger with a larger size of the blur, it stays still relatively small.

In Sec. II the derivation of the search algorithm is presented. In Sec. III, are shown the results of a test of the search algorithm done for a model image. Section IV is for the summary.

## II. METHOD

The given image $g(x,y)$ can be modeled as the convolution of a true image $f(x,y)$ and a blur $h(x,y)$ plus some additive noise $n(x,y)$,

$$g(x,y) = f(x,y) * h(x,y) + n(x,y), \qquad (1)$$

S. Aogaki is with the Department of Information and Communication Sciences, (tel.: +81-75-705-1694, e-mail: aogaki@cc.kyoto-su.ac.jp).
I. Moritani is with the Department of Information and Communication Sciences, (tel.: +81-75-705-1694, e-mail: i654168@cc.kyoto-su.ac.jp).
T. Sugai is with the Department of Information and Communication Sciences, (tel.: +81-75-705-1694, e-mail: sugai@cc.kyoto-su.ac.jp).
F. Takeutchi is with the Department of Computer Sciences, (tel.: +81-75-705-1694, e-mail: takeut@ksuvx0.kyoto-su.ac.jp).
F.M. Toyama is with the Department of Information and Communication Sciences, (tel.: +81-75-705-1898, e-mail: toyama@cc.kyoto-su.ac.jp).

where the position of a pixel is denoted by $x$ and $y$. In this paper we assume an ideal situation where the noise is minute enough to be neglected. In such a case, the observed image is given as the convolution $f(x,y)$ and $h(x,y)$. The $z$-transform $G(u,v)$ of a given image function $g(x,y)$ of its size $M \times N$ is written as

$$G(u,v) = \frac{1}{MN}\sum_{x=0}^{M-1}\sum_{y=0}^{N-1} g(x,y)u^x v^y = F(u,v)H(u,v), \qquad (2)$$

where $u$ and $v$ are complex variables and $F(u,v)$ and $H(u,v)$ are both $z$-transfroms of $f$ and $h$ such as

$$F(u,v) = \frac{1}{m'n'}\sum_{x=0}^{m'-1}\sum_{y=0}^{n'-1} f(x,y)u^x v^y, \qquad (3)$$

$$H(u,v) = \frac{1}{mn}\sum_{x=0}^{m-1}\sum_{y=0}^{n-1} h(x,y)u^x v^y. \qquad (4)$$

For a given $u$, the solutions $v$ of the equation $G(u,v)=0$ are denoted by $\beta^u_i$ ($i=1,2,\cdots,N'$; $N' \leq N-1$). Then $G(u,v)$ of (2) is also expressed as

$$G(u_j,v) = k_j \prod_{i=0}^{N-1}(v-\beta^j_i), \qquad (5)$$

where suffix $j$ means different value of $u$.

The blur function $H$ can also be expanded in the same manner as above;

$$H(u_j,v) = p_j \prod_{i=0}^{n-1}(v-\gamma^j_i), \qquad (6)$$

where $\gamma^j_i$ is the $i$-th solution of $H(u_j,v)=0$. From (2), it follows that $\{\gamma^j\} \subset \{\beta^j\}$.

The RHS of the equation (6) can be expanded as

$$H(u_j,v) = p_j \sum_{y=0}^{n-1} c_y v^y,$$

where $c_y$ is the coefficient of the degree $y$ term, all containing $\gamma^j_0$ through $\gamma^j_{n-1}$ except for $c_{n-1}$ which is just 1. Thus a set of $n$ equations can be constructed as

$$\sum_{x=0}^{m-1} h(x,0)u_j^x = p_j c_0 = (-1)^{n-1} p_j \prod_{i=1}^{n-1}\gamma^j_i$$

$$\sum_{x=0}^{m-1} h(x,1)u_j^x = p_j c_1 = (-1)^{n-2} p_j (\prod_{i=2}^{n-1}\gamma^j_i + \prod_{i=1(i\neq 2)}^{n-1}\gamma^j_i + \cdots + \prod_{i=1}^{n-2}\gamma^j_i) \quad (7)$$

$$\vdots$$

$$\sum_{x=0}^{m-1} h(x,n-1)u_j^x = p_j c_{n-1} = p_j,$$

which can be considered as a set of simultaneous equations with unknowns $h(x,y)$ and $p_j$.

Our aim is to obtain the blur functions $h(x,y)$ from (7). However, (7) has $mn+1$ unknown variables. On the other hand, the number of independent equations in (7) is $n$. Therefore, in order to obtain $h(x,y)$ uniquely, one has to evaluate (7) at $q$ (i.e., $j = 1,\cdots,q$) different values of $u_j$, where $q$ can be taken as the smallest integer that satisfies $q \leq mn/(n-1)$. Thus, one can solve (7) with $j=1,\cdots,q$ for $h(x,y)$ and $p_j$. If (7) has nontrivial solutions for $h(x,y)$ and $p_j$'s then the zero-values $\gamma^j_i$ $(i=1,2,\cdots,n-1)$ are those of a $m \times n$ blur and the solutions $h(x,y)$ is the blur matrix. On the other hand, if (7) has no solution, it means that the zero-values $\gamma^j_i$ $(i=1,2,\cdots,n-1)$ are not of those of a $m \times n$ blur. For any combinations of $n-1$ $\beta^j$'s, one repeats this procedure. The number of the possible combinations is $_{N-1}C_{n-1}$ for a given $M \times N$ image. Once a $m \times n$ blur matrix $h(x,y)$ is found, one can construct the $z$-transform $H(u,v)$ of it. Hence, one can obtain the $z$-transform $F(u,v)$ of a real image $f(x,y)$ simply by $F(u,v) = G(u,v)/H(u,v)$, which allows to restore the true image by the inverse Fourier transform. This makes the image restoration procedure simple. This is an advantage of this search algorithm.

The search algorithm for finding $m \times n$ blur matrix $h(x,y)$ is summarized as follows:
1. Determine the smallest integer $q$ that satisfies $q \leq mn/(n-1)$.
2. Pick up combinations of $n-1$ zero-values $\beta$ from $\beta_i (i=1,2,\cdots,N'; N' \leq N-1)$ of the variable $v$ of the $z$-transform of the given image.
3. Evaluate the $n-1$ zero-values $\beta$ at $q$ different points. Then, solve (5) for the blur $h(x,y)$ and unknown constants $p_j (j=1,\cdots,q)$.
4. If the set of equations give a non-trivial solution, then those $\beta$'s are in reality $\gamma$'s which are solutions of $H(u_j,v)=0$.
5. Find the nontrivial solutions by repeating the procedures 1. ~ 3 at most $_{N-1}C_{n-1}$ times until the condition 4. is met.
6. Restore the true image $f(x,y)$ by removing the blur using $F(u,v) = G(u,v)/H(u,v)$.

When one chooses $q$ different values for $u_j$'s, it is advisable to select values which are not very different each other. In that case, if the $i$-th solution $\beta^j_i$ is in reality a $\gamma^j$, the chance that $\beta^{j'}_i$ at another value $u_{j'}$ which is close to $\beta^j_i$ is also a solution $\gamma^{j'}$.

We have constructed the algorithm in terms of the zero-values of the variable $v$. Of course we may use the zero-values $\alpha_i$ $(i=1,2,\cdots,m-1)$ of $u$ instead of $v$ in constructing the same algorithm. Such a version is simply obtained by the replacements $N \to M$, $n \to m$, $\beta_i \to \alpha_i$, and $u_j \to v_j$.

III. TEST OF THE SEARCH ALGORITHM

In the preceding section we have discussed a search algorithm for finding a single blur convolved in a given image. In this section is presented one of the results of tests of image restoration that have been performed using the present search algorithm.

Fig.1 shows the test images that are the same as those used in [2]. Fig. 1(a) shows a $40 \times 40$ model image that can be regarded as a true image. Figs. 1(b), 1(c) and 1(d) are blurs of sizes $2 \times 2$, $2 \times 3$ and $3 \times 3$, respectively. We convolve the three blurs into the true image. Figure 1(e) shows the convolved image. The size of the convolved image is $44 \times 45$.

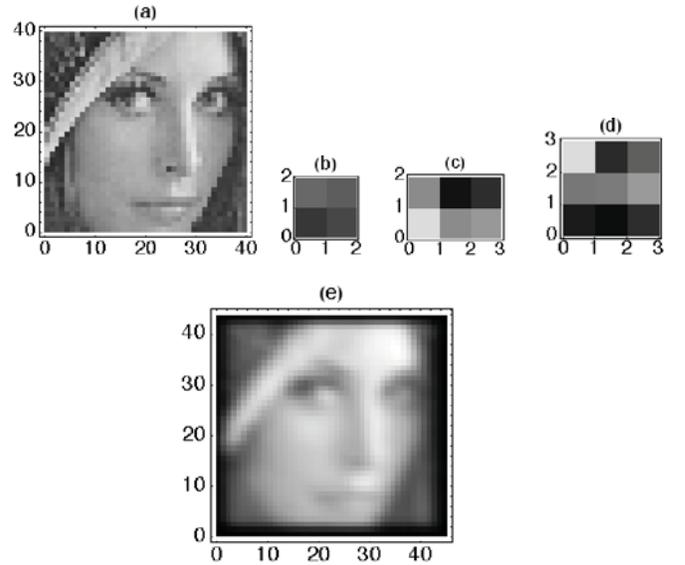

Fig. 1. (a): True image of size $40 \times 40$ that we took from [3]. (b): Blur image of size $2 \times 2$. (c): Blur image of size $2 \times 3$. (d): Blur image of size $3 \times 3$. (e): Image by convolving the three blurs of (b), (c), and (d) into (a). The image size is $44 \times 45$.

We test how the search algorithm represented by (7) works in finding each single blur convolved in the true image. To choose $u_j$'s we imposed $|u_j|=1$, and changed only its phase.

We used *Mathematica* to obtain the zero-values $\beta_i$ of $G(u,v)$ solving the simultaneous equation (5) with enough precision. The parameter $q$ is 4, 3, and 5 for each blur of Fig. 1(b), 1(c), and 1(d). Fig.2 shows the results of the test of the search algorithm. Fig. 2(a) shows the restored image by removing the $2 \times 2$ blur that has been detected by searching the $2 \times 2$ blur in Fig. 1(e). Fig. 2(b) shows the restored image by removing the $2 \times 3$ blur that has been detected by searching the $2 \times 3$ blur in Fig. 2(a). Finally, Fig. 2(c) shows the restored image by removing the $3 \times 3$ blur that has been detected by searching the $3 \times 3$ blur in Fig. 2(b). In obtaining the final restored image of Fig. 2(c) we applied the search algorithm three times. In each search, the algorithm worked very well for finding each single blur image of different size. All calculation in this test has been performed using *Mathematica*. In this way we could verify that the search algorithm certainly works well in finding a single blur convoluted in a given image. This test is just one of many tests that we have carried out.

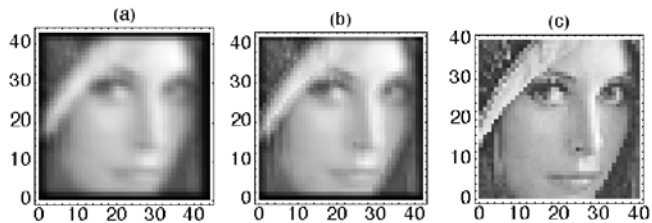

Fig. 2. Restored images by removing three blurs that were searched for by the search algorithm. (a): restored image by removing the $2 \times 2$ blur from the image of Fig. 1(e); (b): restored image by removing the $2 \times 3$ blur from the image of (a); (c): restored image by removing the $3 \times 3$ blur from the image of (b).

## III. SUMMARY

A simple search algorithm for finding blurs convolved in a given image is presented. The search algorithm is given as simultaneous equations for a blur matrix elements. The algorithm is for finding a single blur of a specified size. Once a blur of a given size is detected, and the matrix elements are obtained, one can easily reconstruct the unblurred image from the *z*-transform of the obtained blur. The advantage of this method is that it can be extended easily to a larger size of blurs.

We experimentally tested the algorithm by using a test image to examine how it works for finding a blur convolved in an original image. We have verified that the algorithm works, in a practical processing time, very well for blurs of small sizes convolved in a middle size image. This method would be very useful in making use of the LB blind deconvolution.

ACKNOWLEDGMENT

This work was supported in part by the Science Research Promotion from the Promotion and Mutual Aid Corporation for private Schools of Japan, and a grant from Institute for Comprehensive Research, Kyoto Sangyo University.